\documentclass[sigconf]{acmart}
\AtBeginDocument{%
  }
\newcommand*\circled[1]{\tikz[baseline=(char.base)]{
            \node[shape=circle,draw,inner sep=1pt] (char) {#1};}}
\newcommand{\tap} {\texttt{TAPAS}\xspace}
\usepackage{balance}
\usepackage{natbib}
\usepackage{graphicx}
\usepackage{textcomp}
\usepackage{xcolor}
\usepackage{microtype}
\usepackage{graphicx}
\usepackage{amsmath}
\usepackage{hyperref}
\usepackage{listings}
\usepackage{subfigure}
\usepackage{booktabs}
\usepackage{float}
\usepackage{listings}
\usepackage{multirow}
\usepackage{algorithm}
\usepackage[noend]{algpseudocode}
\usepackage{xspace}
\usepackage{tikz}
\usepackage{subcaption}

\lstdefinestyle{Python}{
    language        = Python,
    basicstyle      = \small\ttfamily, 
    deletekeywords={import},
    frame=tb,
    captionpos=t,                    
    escapeinside={(*}{*)},
    columns=fixed ,
    keepspaces=true,
     breaklines=true, 
     belowcaptionskip = 5pt,
}
\setcopyright{acmlicensed}
\copyrightyear{2025}
\acmYear{2025}
\acmDOI{3754598.3754677}
\acmConference[ICPP '25]{54th International Conference on Parallel Processing}{September 08--11,
  2025}{San Diego, CA}
\acmISBN{979-8-4007-2074-1}

\acmSubmissionID{pap458s1}

\begin{document}

\title{TAPAS: Fast and Automatic Derivation of Tensor Parallel Strategies for Large Neural Networks}

\author{Ziji Shi}
\authornote{Work done during internship at Alibaba Cloud.}
\orcid{0000-0001-9398-6507}
\affiliation{%
  \institution{National University of Singapore}
  \institution{Alibaba Group}
  \country{Singapore}
}
\email{ziji.shi@u.nus.edu}

\author{Le Jiang}
\orcid{0009-0002-9941-2322}
\affiliation{%
  \institution{Alibaba Group}
  \city{Hangzhou}
  \country{China}}
\email{jiangle.jl@alibaba-inc.com}

\author{Ang Wang}
\orcid{0009-0007-2650-0504}
\affiliation{%
  \institution{Alibaba Group}
  \city{Hangzhou}
  \country{China}}
\email{wangang.wa@alibaba-inc.com}

\author{Jie Zhang}
\orcid{0009-0003-5085-2535}
\affiliation{%
  \institution{Alibaba Group}
  \city{Hangzhou}
  \country{China}}
\email{wanglin.zj@alibaba-inc.com}

\author{Chencan Wu}
\orcid{0009-0005-1398-8235}
\affiliation{%
  \institution{Alibaba Group}
  \city{Hangzhou}
  \country{China}}
\email{danguge@buaa.edu.cn}

\author{Yong Li}
\orcid{0000-0001-9072-3170}
\affiliation{%
  \institution{Alibaba Group}
  \city{Hangzhou}
  \country{China}}
\email{jiufeng.ly@alibaba-inc.com}

\author{Xiaokui Xiao}
\orcid{0000-0003-0914-4580}
\affiliation{%
  \institution{National University of Singapore}
  \country{Singapore}}
\email{xkxiao@nus.edu.sg}

\author{Wei Lin}
\orcid{0000-0002-3003-0150}
\affiliation{%
  \institution{Alibaba Group}
  \city{Hangzhou}
  \country{China}}
\email{weilin.lw@alibaba-inc.com}

\author{Jialin Li}
\orcid{0000-0003-3530-7662}
\affiliation{%
  \institution{National University of Singapore}
  \country{Singapore}}
\email{lijl@comp.nus.edu.sg}


\begin{abstract}
Tensor parallelism is an essential technique for distributed training of large neural networks. However, automatically determining an optimal tensor parallel strategy is challenging due to the gigantic search space, which grows exponentially with model size and tensor dimension. This prohibits the adoption of auto-parallel systems on larger models. 

We observe that neural networks usually contain repeated substructures, and build an automatic parallelism framework named TAPAS that eliminates redundant search efforts. TAPAS employs a divide-and-conquer approach that efficiently folds the search space by identifying those unique substructures. As a result, it runs at sub-linear complexity concerning the model size, making it a scalable solution for training large-scale networks. Our evaluations demonstrate that TAPAS outperforms the state-of-the-art automatic parallelism frameworks by up to $160\times$ in search speed on a wide range of models, and the performance of derived strategies is competitive or even better compared with the expert-engineered Megatron-LM library.
\end{abstract}




\maketitle
\section{Introduction}
Model scaling have been the cornerstones in neural network advancements in recent years, resulting in many powerful and gigantic models.
Researchers have observed it that model can perform better by increasing the number of parameter, training on larger datasets, and supplying more compute~\cite{Kaplan2020ScalingModels}. 
This has led to the advancements of many powerful models like DeepSeek-V3~\cite{liu2024deepseek}, Llama~\cite{touvron2023llama}, and GPT ~\cite{Brown2020LanguageLearners}.
However, the advancement of memory capacity in AI accelerators has not kept in pace.
Over the last decade, the memory capacity of Nvidia GPUs has only increased by 16 times (from K20 to H100) to reach 80GB, whereas the size of neural networks has expanded by 30,000 times (from AlexNet~\cite{krizhevsky2012imagenet} to GPT-4o).
To address this problem, researchers propose model parallelism, where model weights and optimizer stats are sharded across multiple machines during distributed training.

However, several challenges on designing training strategies surface as model size scales up.
Firstly, manually specifying the optimal parallel strategy is becoming increasingly difficult.
While large-language models (LLMs) are popular, there exists many non-transformer-based models for targeted applications. 
For instance, U-Net\cite{ronneberger2015u} is a "U"-shaped convolutional neural network (CNN) used for image segmentation tasks, particularly in medical imaging.
Recommendation system usually adopts a two-tower model architecture\cite{covington2016deep} for mapping user and item features respectively, where each tower has a different design. 
Large-scale classification models consist of a feature extraction module and a classification module that scales with the number of targets. 
It is challenging to design a simple strategy that fits all.
On top of it, the optimal parallel strategy often requires a combination of multiple parallelization techniques and in-depth knowledge about the underlying system.
Therefore, relying solely on expert knowledge to design such strategies is becoming intractable and error-prone.  

Furthermore, the number of possible strategies in tensor parallelism grows exponentially with model size.
The total number of possible tensor parallel strategies is determined by the Cartesian product of number of tensors and their orders across all tensors in the model. 
Since recent models can contain thousands of tensors, the total number of possible strategies quickly grows beyond what is feasible to explore exhaustively.
For instance, DeepSeek-V3 has 91997 weight tensors, and Llama3.1-405B has 1140 weight tensors.
Assuming each tensor has at least three dimensions (batch, sequence, hidden), there are 778.6 trillion and 1.5 billions possible tensor parallel strategies respectively.
Therefore, performing exhaustive search over the entire search space can be prohibitively expensive in practice, limiting the adoption of existing automatic model parallel systems.

On top of this, strategy validation becomes computationally expensive as the number of candidate strategies grows. 
To ensure the new parallel strategy is mathematically equivalent as the original model, each strategy needs to be validated. 
As the number of candidate strategies scales up, the current dry-run or randomized-testing-based validation method can be a performance bottleneck, and we must design better validation scheme to early-stop on infeasible strategies. 

Due to the challenges above, existing state-of-the-art solutions for training auto-parallelism usually suffer from prohibitively long search time.
Concretely, Alpa takes 5.8 hours to search on a GShardMoE-2.4B model, yet it is projected to take more than 6 million hours to derive the strategy for a 240 billion counterpart.
This scalability limitation makes it challenging to employ auto-parallel systems on large neural network models.  
\begin{figure}[t]
\centering
\includegraphics[width=\linewidth]{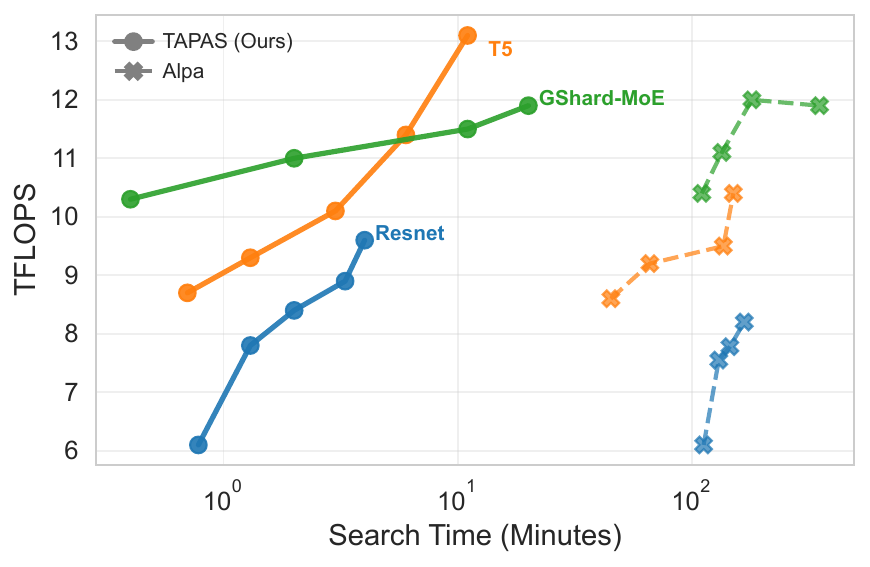}
\caption{Search time budget vs. training throughput.}
\label{fig:teaser}
\end{figure}

\begin{table*}
	\centering
	\begin{tabular}{c|c|c|c|c}
		\hline  Framework & Search Space & Search Algorithm & Strategy Validation & Overall\\ 
		\hline \textbf{FlexFlow} & $N(4E, 4V)$ & $O(B)$ & $O(V+E)$ & $O(BV+BE)$\\
		\hline \textbf{Alpa} & $N(kE, kV) $ & \begin{tabular}{@{}c@{}} Inter-Op: $O(V^2L)$ \\ Intra-Op: $O(E(V+E))$ \end{tabular} & $O(V+E)$ &$O(V^2L(V+E^2))$\\
		\hline \textbf{\tap} & $N(\frac{E}{2CL}, \frac{V}{2CL})$ & $O(\frac{E+V}{L})$ & $O(\frac{E}{L})$ & $O(\frac{E+V}{L})$\\
		\hline
	\end{tabular} 
	\caption{Complexities of selected auto parallel frameworks. $E$ and $V$ are the number of connections and number of vertices in the computational graph respectively, $L$ is the number of layers, $C$ is the frequency of repeated substructure.}
	\label{tab:algo-complexities}
\end{table*}

In this work, we propose a novel automatic tensor parallel system that significantly reduces the strategy search time without compromising strategy quality, as shown in Figure~\ref{fig:teaser}.
It is based on the key observation that repeated substructures (i.e., reused layers/operator groups) are commonplace in network architectures. 
These repeated substructures can be exploited to effectively \emph{fold} the search space to increase strategy derivation speed.
We further show that a parallel strategy, when applied to two similar layers in different parts of the model, exhibits similar resource requirements.
This is because the same layers share equivalent communication, computation, and memory access patterns.
A direct corollary is that a single parallel schedule, once derived, can be reused for all repeating layers without loss of efficiency.

Guided by this observation, our approach folds the search space by identifying the set of \emph{unique sub-computational graphs}, each representing a unique network.
Thereafter, we restrict the search space from the entire computational graph to the set of unique subgraphs, resulting in an exponential decrease in search complexity.
To further accelerate the strategy validation process, we adopt an early-stopping framework on the candidate strategy.  
After that, the remaining challenge is to ensure the optimized subgraphs will combine to form a valid solution.
We employ static analysis to verify that the derived sub-strategies are valid and compatible, and that the final parallel strategies maintain mathematical equivalence with the original model.
In the end, our system selects the best strategy using a communication-based cost model and reconstructs the parallelized computational graph.

We present \tap (\underline{T}ensor \underline{A}uto \underline{Pa}ralleli\underline{s}ation), an automatic parallel framework that efficiently derives tensor parallel strategy for a given neural network. 
\tap requires no expert annotations, and achieves $20-160\times$ search time speedup over the state-of-the-art auto-parallel framework Alpa ~\cite{Zheng2022Alpa:Learning}.
With the growing size of foundation models, \tap proves to be a scalable option. 
We further demonstrate that \tap can identify a tensor parallel strategy with comparable performance as expert-designed solutions like Megatron-LM ~\cite{Shoeybi2019MegatronLM} or DeepSpeed\cite{rasley2020deepspeed}.


We summarize our contribution as the following:
\begin{itemize}
    \item We identify the problem that automatic searching for a parallel strategy can be slow and inefficient, formulate the prior works under one view, and provide a grounded analysis of the complexity.

    \item Observing that parallel strategies are similar for similar layers/blocks, we propose a computational graph pruning method that efficiently folds the search space into a limited set of subgraphs. 
    
    \item We design and implement \tap, an automatic tensor parallel framework. \tap can discover tensor parallel strategies with better or matching performance than the State-of-the-Art strategies, while being two orders of magnitude faster in searching. 
\end{itemize}

\section{Related Work}
In this section, we outline the current research landscape for model parallelism and automatic parallelism.



\subsection{Model Parallelism}

Tensor parallelism and pipeline parallelism are two commonly used approaches for model parallel training.

\subsubsection{Pipeline Parallelism (PP)}
Pipeline parallelism divides a model by layers and assigns each group of layers to a separate device~\cite{Huang2019GPipe:Parallelism, Narayanan2019Pipedream:Training, Fan2021DAPPLE:Models, qi2024pipeline}. A training batch is split into micro-batches so that different devices can compute in an overlapped fashion. The efficiency of PP is determined by the bubble time, where devices are idle due to unsatisfied dependency. To keep bubble time low, each pipeline stage should use similar amounts of memory and computation so that workloads remain balanced\cite{qi2024pipeline}. In practice, the heterogeneity in models or layer interdependency may limit even pipeline splitting.
For example, large classification models often end with a heavy fully-connected output layer, and encoder–decoder models can suffer from cross-attention dependencies between encoder and decoder stages.

\subsubsection{Tensor Parallelism (TP)}
Tensor parallelism, also called tensor sharding, partitions each layer at the tensor level and places each shard on a different device~\cite{shazeer2018mesh,Shoeybi2019MegatronLM,Narayanan2021EfficientMegatron-LM}. When a full weight or activation tensor is required, TP uses collective operations such as all-gather or all-reduce to reassemble the full tensor from its shards. 

Compared to PP, TP typically incurs higher communication overhead because all devices in the group must exchange data. In addition, the space of possible TP configurations is much larger than for PP, making it more challenging to find an optimal partitioning.

\subsection{Automatic Parallelism}
With rich options for parallel strategy, it has become difficult to determine which one to use for distributed training. 
Some existing works focus on building customized training systems for specific models like embedding model~\cite{Miao2022HET, Zhang2024CAFE}, Generative Adversarial Networks~\cite{shi2024paragan}, Mixture-of-Expert models~\cite{he2022fastermoe, Nie2023FlexMoE}, and Graph Neural Networks ~\cite{Zhang2023Lotan, Wang2022DistributedGNN}. 
Those systems base their optimizations on model-specific characteristics and thus cannot generalize to new models.
Automatic parallelism is a recent line of research on automatically selecting parallel strategies for distributed training with minimal user intervention. 

Because of the vast search space of parallel strategies, existing works on automatic parallelism either rely on user annotation or brute-force searches over the all possible candidates. 

\subsubsection{Directive-based approaches}
Directive-based automatic parallelism relies on expert annotations to derive parallel strategies.
Those annotations are usually bound to specific model dimensions.
For example, Mesh TensorFlow~\cite{shazeer2018mesh} infers the operator partitioning scheme based on user-defined directives to scale single-device programs.
Whale~\cite{whale2022} allows for incorporating user annotation to perform semi-auto parallelisation for large models and introduces a hardware-aware load balance algorithm.
However, directive-based automatic parallelism approaches require users to have a deep understanding of both the system and the model, and the hard-coded user annotations may not be transferable when either the model or system changes.

\subsubsection{Search-based approaches}
Recent work has proposed fully automatic approaches based on search algorithms to optimize distributed DNN training.
For example, Tofu~\cite{Wang2019SupportingPartitioning} uses a recursive search algorithm to derive a communication-efficient schedule for the CNN and RNN models, but it does not generalize to Transformer-based architectures with many dense MatMul operators.
Leveraging the discrete nature of candidate spaces, Flexflow~\cite{Jia2018BeyondNetworks} uses Markov-Chain Monte Carlo search to find the best parallel strategy for DNN models. 
While this approach works well for small-scale neural networks, it cannot scale well on large neural networks without properly engineered heuristics. 
Alpa~\cite{Zheng2022Alpa:Learning} adopts a two-level optimization approach: it uses an inter-operator optimization pass to cluster operators, and a secondary intra-operator optimization pass to find tensor parallel strategies within the cluster. 
Unity~\cite{Unger2022UnityThe} represents both parallelisation and algebraic transformations in a unified manner, and uses a hierarchical search algorithm to identify an optimized sequence of graph substitutions. 

\subsubsection{Challenge}
While search-based approaches have demonstrated promising results, they encounter a major obstacle: the exponential growth of the search space leads to a prohibitively long search time. Specifically, for neural networks, each N-dimensional tensor offers N+1 possible strategies: not sharding, or sharding along the N-th dimension. Consequently, for a neural network represented as $G(E,V)$ with $V$ tensors, the number of possible tensor parallel strategies can reach up to $(N+1)^{V}$. As a result, identifying an optimal sharding strategy is beyond the capabilities of polynomial time algorithms, underscoring the critical challenge of managing the vast search space efficiently.

This exponential increase in strategy space has led to impractical search time to derive parallel schedules for large models, which we will later show in \autoref{sec:e2e-search}. The search speed problem has become an emerging bottleneck in training foundation models.

\section{Approach}
Can we accelerate the derivation by leveraging the insights from the model architecture?
In this section, we begin by introducing two common patterns of model scaling (by width and by depth), and the challenges associated with the existing approaches in handling these cases.
We then formulate the problem of finding the parallel strategy.
In the end, we propose to use graph pruning to reduce the search space.

\subsection{Motivating Examples}


We review common model scaling techniques, and propose that these techniques can be grouped into two major categories:
\textbf{scaling on the width} by increasing the dimension of layers (e.g., adding the number of classes, adding attention heads, or increasing the convolutional channels), or \textbf{scaling on the depth} by increasing the number of layers.

We give two concrete examples of model scaling below.

In the context of e-commerce, a wide range of merchandise exists, potentially numbering in millions to billions. 
Consequently, a product image classification model such as ResNet~\cite{He2016DeepRecognition} requires an exceptionally wide fully connected (FC) layer to classify it. When the number of classes reaches 100,000, the FC layer will comprise 205M floating point numbers, significantly outweighing the feature extraction module, which stands at a modest 24M parameters.


In the scaling-on-depth scenario, we examine models based on the transformer architecture. 
Currently, most large language models employ the transformer layer~\cite{Vaswani2017AttentionNeed}, which includes an attention module followed by a feedforward network.
In the quest for scale, dense transformer models typically stack more transformer layers on top of each other ~\cite{Devlin2019BERT:Understanding, widenet, gpt3, Dosovitskiy2020AnScale}, driven by the observation that larger models usually perform better ~\cite{Lepikhin2020GShard:Sharding, tay2021scale, wei2022emergent}.
Because of the uniformity in model architecture, there exists potential for reusing the sharding pattern discovered for one layer across all transformer layers~\cite{Narayanan2021EfficientMegatron-LM}.

In summary, it is increasing difficult to build tailored training systems in response to exponentially growing model sizes.
How to \textit{automatically} and \textit{quickly} devise a parallel strategy to train these models? 
We formulate the problem of the automatic derivation of parallel strategy as a graph transformation problem, and present the challenges in the next section. 


\subsection{Problem Formulation}
We formulate the problem using the graph representation of neural networks. 
All neural networks can be represented as a directed acyclic graph $G(E,V)$ comprised of $L$ layers.
The set of vertices $V$ represents the operators, and the set of edges $E$ represents the data flow from producer to consumer operators.
During the forward pass, an edge represents an activation tensor, while in the backward phase, an edge represents a gradient or error tensor.
A layer $L_i \in L$ could consist one or more nodes. 
The training cluster is modeled as $S(m,n)$ where $m$ is the number of worker nodes, and $n$ is the number of accelerators per worker node.
A parallel strategy $P_G$ is a graph transformation on $G$ that preserves the mathematical equivalence.
The goal is to find an optimal parallel strategy ${P_G}^*$ such that: given any input, the output is equivalent and the training throughput is maximized.



The end-to-end duration to produce an optimal schedule is a critical metric for an auto-parallel system.
We identify three main factors that contribute to this overall completion time: the size of the search space, the time complexity of the searching algorithm, and the speed of the evaluation method.
Out of them, the search space is the most important factor as it determines the input size of the other two factors. 



\textbf{Observation \#1: repeated subgraphs commonly exist in large models.}
As we see earlier, a major challenge faced by auto-parallel systems is the search space explosion problem.
Our key observation is that both scaling by width or depth techniques start with a \textit{base subgraph}, i.e., a group of layers or operators, and expand from it. 
For instance, large-scale pre-trained language models such as BERT~\cite{Devlin2019BERT:Understanding} and T5~\cite{Raffel2020ExploringTransformer} consist of tens of transformer layers, and multi-class object classification networks like ResNet-50~\cite{He2016DeepRecognition} are made of repeated convolutional layers. 
In DeepSeek-v3/R1 model, the basic components are MLP, self-attention, and MoE layers. 
This indicates that we can save much effort by restricting the search on the subgraphs instead of the whole graph, where a lot of search efforts are wasted on identical structures. 

\textbf{Observation \#2: the performance of repeated subgraphs is identical across the model.}
Furthermore, by analyzing expert-engineered parallel schedules~\cite{Narayanan2021EfficientMegatron-LM, Rajbhandari2020Zero:Models, Ren2021ZeRO-offload:Training}, we observe that parallel schedules are primarily identical for the same type of layers. 
The underlying reason is the same layers share the same amount of communication, computation, and memory access patterns.
Therefore, their performance should also be similar. 

This has motivated us to explore the possibilities of reusing the parallel schedules discovered for the same layer to save search effort. Specifically, we first identify the unique subgraphs through pattern matching, derive the parallel strategy for each subgraph, then apply the same strategy to the rest of the subgraphs. 

Another challenge we face is the complexity of parallel implementations of operators. Given the huge variety of operators, with each of them having multiple possible sharding implementations, how to flexibly represent all possible combinations, and ensure the final implementation is correct?
To answer this, we adopt an \textit{enumerate-then-validate} approach by representing all possible strategy combinations using a decision tree-like structure, then validating the tree and performing early stopping if necessary. 
By doing so, we avoid running the strategies and saved efforts on incorrect strategies. 
\section{Design and Implementation}
\begin{figure}
\centering
\includegraphics[width=\linewidth]{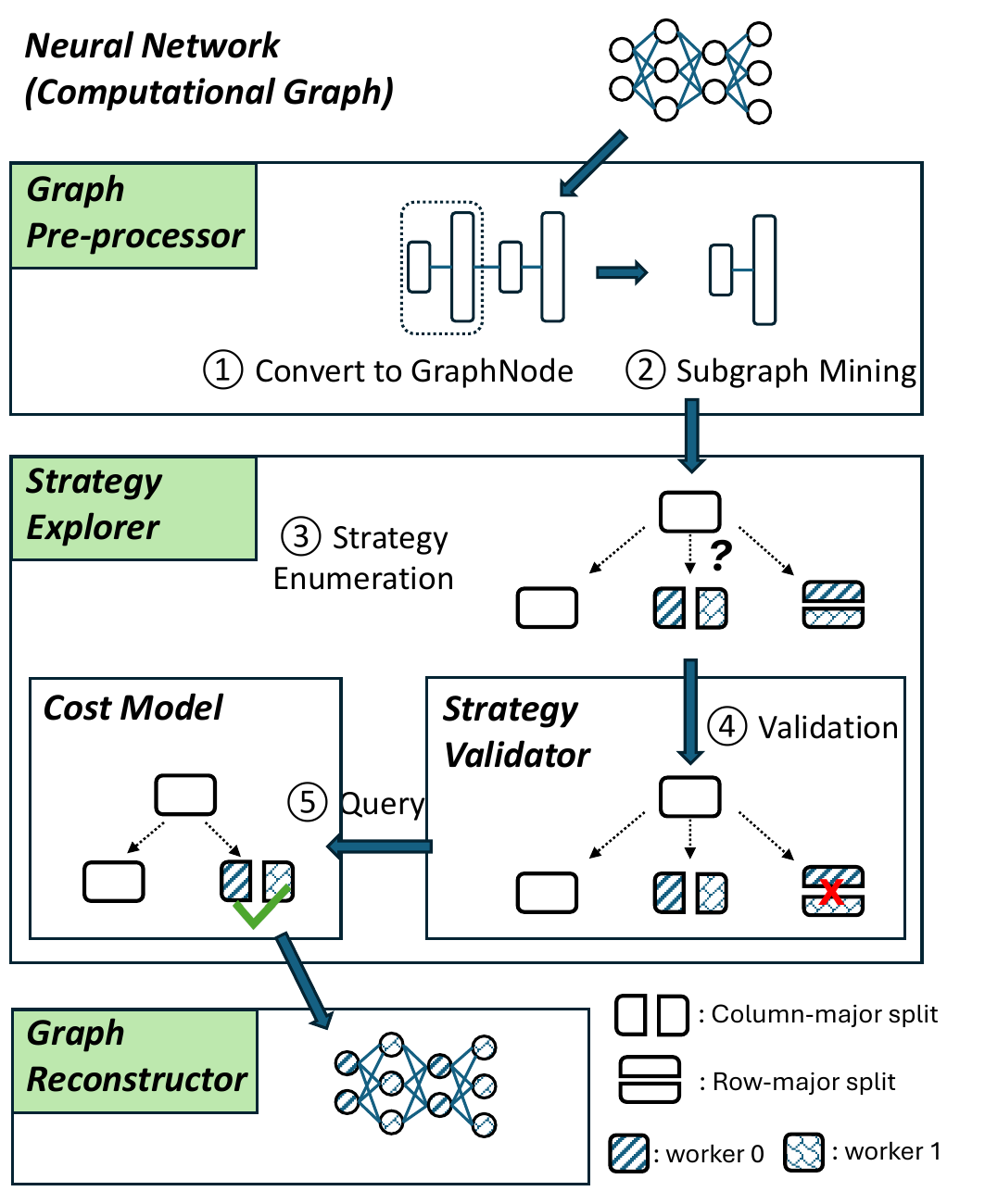}
\caption{TAPAS system architecture. }
\label{fig:system-overview}
\end{figure}

\subsection{Overview}
Based on the insights, we design TAPAS, a tensor auto-parallel framework focusing on the strategy derivation speed. 
\tap significantly reduces redundant search efforts by shrinking the search space at different levels without compromising the strategy quality.

As depicted in \autoref{fig:system-overview}, given any neural network, \tap\ first converts the graph into groups of operators named GraphNodes (Step \circled{1}). 
\tap then performs subgraph mining, restricting the search space from the whole graph to the set of unique subgraphs (Step \circled{2}).
After finding the unique subgraphs, \tap enters the Strategy Exploration phase by enumerating all possible parallel strategies for each unique subgraph based on the sharding patterns (Step \circled{3}). 
After that, it validates each strategy to ensure the new computational graph can be correctly reconstructed later(Step \circled{4}). 
At the end of the Strategy Exploration phase, all remaining candidate strategies are evaluated using the cost model (Step \circled{5}).
In the end, \tap takes the best parallel strategy and reconstructs it back to the computational graph for execution on DL framework backends like TensorFlow.

\subsection{Intermediate Representation}\label{sec:ir}
\tap\ preprocesses the computational graph into Intermediate Representations (IRs) to facilitate the derivation of parallel strategies.
Compared to other deep learning IRs like MLIR HLO~\cite{mlir}, \tap\ IR groups operators that are collectively used together. 
Each group of operators(\textit{GraphNodes}) is associated with a set of possible sharding implementations (\textit{ShardingPatterns}) expressed using the Split-Replica-Communication (SRC) expression.
Once the sharding rules are determined, the associated costs are also decided, which are then used to evaluate the strategies. 





\begin{figure}[h]
\centering
\includegraphics[width=\linewidth]{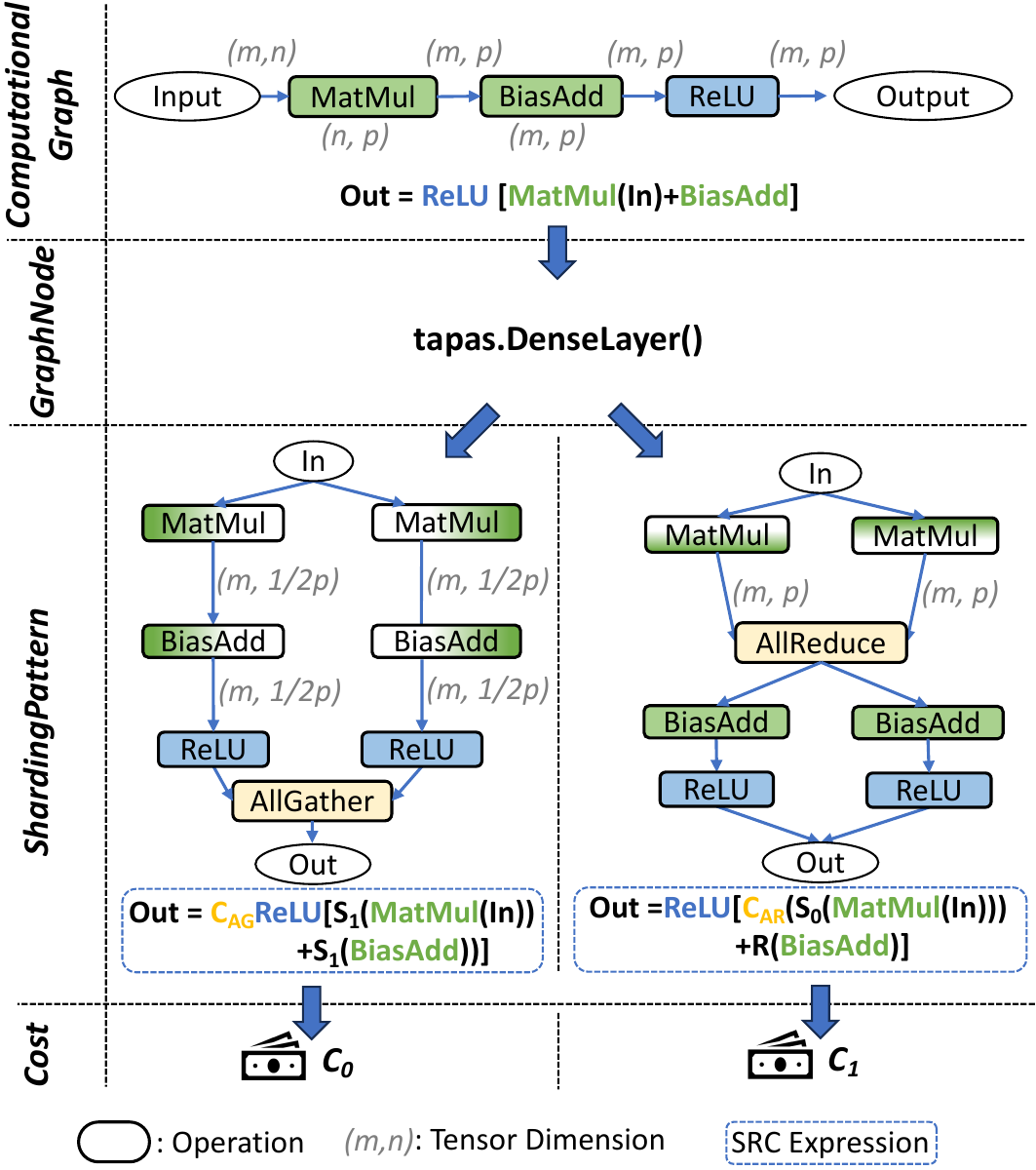}
\caption{Overview of graph transformation for dense layer.}
\label{fig:tapas_ir}
\end{figure}

\paragraph{GraphNode.}
\textit{GraphNode} is the basic unit for deriving the parallel strategy. 
It is a container of operators collectively used together.
GraphNode is introduced because \textit{the sharding decision is interrelated within a layer}: a decision on the previous tensor will affect the tensor after. 
In the GraphNode representation, \tap tracks the input/output shape and the split axis of each tensor.
For instance, in \autoref{fig:tapas_ir}, a dense layer can be a GraphNode, which consists of a matrix multiplication (MatMul) op, an addition (BiasAdd) op, and an activation (ReLU) op that has no weight. 


\paragraph{ShardingPattern.}
A \textit{ShardingPatterns} is a possible parallelised implementation of a GraphNode. 
For instance, a 2D matrix weight can be split on either dimension or replicated.
Therefore, a GraphNode can have multiple valid ShardingPatterns.
For each GraphNode, \tap defines its sharding patterns using the SRC expression (\autoref{sec:src}) to separate the definition from implementation. 
Once the implementations are decided, \tap can estimate its cost based on the communication pattern and tensor size.

\paragraph{Parallel strategy.}
A parallel strategy is a parallelized implementation of the original graph. 
It is constructed by substituting the GraphNodes with their parallel implementations. 
The cost of a parallel strategy is calculated as the sum of the costs of all constituting sharding patterns.

\subsubsection{Split-Replica-Communication Expression.}\label{sec:src}


\tap\ represents all parallel strategies using an abstraction called \textit{Split- Replica- Communication (SRC)}:

\textbf{Split.}
\emph{Split} means sharding the tensor on a target axis, after which different devices store different partitions. 
For example, $S_0(T)$ means sharding tensor $T$ on the first axis.
Under this view, data parallelism is just a special case for tensor parallelism where the tensor shards on the batch dimension.

\textbf{Replica.}
\emph{Replica} ($R(T)$) means to replicate the tensor $T$ on different devices. 
For instance, in data parallelism, the weight tensors are replicated while the input tensors are shared. 

\textbf{Communication.} 
Additional communication operators may be needed to combine the partial results. 
For instance, AllReduce ($C_{AR}$) is needed in data parallel to aggregate gradients, while AllGather ($C_{AG}$) is required to exchange partial values after a split operation. 
It is worth noting that a parallel schedule may not include all three strategies.

Under the SRC expression, an operation 
\[ Y = Op (A, B)\]

can be expressed using SRC expression as:

\[Y = C(Op(S/R(A), S/R(B))  \]


With the SRC expression, we can define general tensor parallel rules for each operator as ShardingPattern.
After the best parallel strategy is selected, sharding patterns will be materialized (reconstructed) into a parallelized graph.



The benefits of using SRC is threefold. 
Firstly, compared to the abstractions in other works ~\cite{Wang2019SupportingPartitioning, Xu2021GSPMD:Graphs, Unger2022UnityThe}, SRC reduces the amount of effort necessary to define parallel implementations for new operators/layers. 
Secondly, having known the tensor shape, SRC can enable symbolic shape checks to validate the parallel strategies.
This is crucial because most consecutive sharding patterns are not compatible, and the search can backtrack earlier if a strategy is deemed invalid.
Last but not least, SRC can be used to express complex patterns through nested expression.

\subsection{Subgraph Mining}\label{sec:pruning}
Given a GraphNode graph, \tap searches for the set of unique subgraphs within it. In order to find significant subgraphs to avoid excessive search, we can control the minimal threshold for subgraph occurrence and subgraph size using \textit{minSupport} and \textit{minSize} respectively.

\begin{algorithm}
\caption{Apriori Frequent Subgraph Search}\label{algo:graph-prune}
\begin{algorithmic}[1]
\Require G(V, E), minSupport, minSize
\Ensure Set of subgraphs with at least minSize nodes
\State Initialize $F \leftarrow$ empty list \Comment{Initialize frequent subgraphs}
\State Initialize $C \leftarrow$ list of V \Comment{Initialize candidate set}
\For{each subgraph $s$ in $C$}
    \If{Count($s$ in G) $\geq$ minSupport}
        \State Add $s$ to $F$
    \EndIf
\EndFor
\State $C \leftarrow$ subgraphs in $F$ \Comment{Candidates for merging}
\For{$k = 2$ to $|V|$ }
    \State $C_{k} \leftarrow$ Merge subgraphs in $F_{k-1}$ that share edges
    \For{each subgraph $s$ in $C_{k}$}
        \If{Count($s$ in G) $\geq$ minSupport and $|s|$ $\geq$ minSize}
            \State Add $s$ to $F_{k}$ \Comment{Add subgraph of size $k$}
        \EndIf
    \EndFor
    \If{$F_{k}$ is empty}
        \State Break \Comment{Stop if no more freq. subgraph of size $k$}
    \EndIf
    \State $C \leftarrow$ subgraphs in $F_{k}$ 
\EndFor 
\Return $F$
\end{algorithmic}
\end{algorithm}


Our subgraph mining algorithm is inspired by the Apriori algorithm in frequent itemset mining. 
It starts by treating each node as a single-node subgraph and considers them as initial candidates. 
Each candidate subgraph is then counted in the graph for frequency, and if its frequency surpasses the minimum support threshold and its size meets a minimum number of nodes criterion, it is considered a significant subgraph and added to the output list (line 3-6).
It then iteratively expands these candidates by merging the existing frequent subgraphs that share a common edge, thereby generating larger candidate subgraphs (line 8).
The algorithm repeats this process until it finds no new frequent subgraphs. 
The final output is a list of all identified frequent subgraphs that contain at least a specified number of nodes.
As \texttt{minSupport} is directly linked to the frequency of subgraph, we set to be the number of layers.

\subsection{Parallel Strategy Exploration}
\label{sec:plan-gen}
 \begin{figure}[h]
    \centering
    \includegraphics[width=\linewidth]{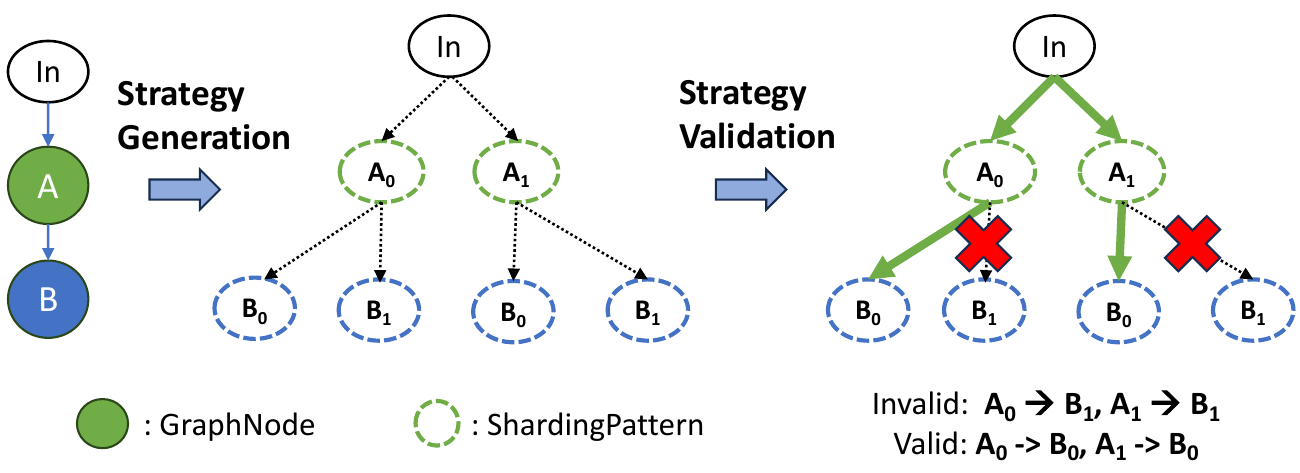}
    \caption{Strategy generation and validation. }
    \label{fig:pse}
\end{figure}

After finding the frequent subgraphs, \tap\ moves on to the Strategy Exploration phase, as illustrated in \autoref{fig:pse}. 
Recall that each subgraph consists of one or more GraphNodes, with each of them potentially having multiple ShardingPatterns. Therefore, this step is accomplished by enumerating all possible combinations of ShardingPatterns within each subgraph and subsequently interlinking them. 

However, simply connecting these ShardingPatterns does not guarantee a functional parallel strategy. Because of possible issues such as shape mismatch between subgraphs, not all combinations yield \textit{valid} parallel strategies. To address this, \tap\ employs the SRC to conduct a symbolic shape check on the ShardingPatterns.

The symbolic shape check involves analyzing the shapes of tensors in the computational graph and ensuring that the operations are compatible across ShardingPatterns. 
The shape propagation is made possible with GraphNodes (which records the shape information of the original tensor) and ShardingPattern (which tracks the transformation rules). 
As shown in \autoref{fig:pse}, the strategy is valid only if every \textit{pair} of consecutive ShardingPattern is valid; otherwise, it is deemed invalid and we can early stop it without exploring this strategy to the fullest.
The compatibility test is particularly crucial when tensors are divided across different devices, and we observe that the vast majority of the strategies are invalid.

Following the validation of strategies, \tap\ constructs the successful parallel strategies using a Breadth-First-Search (BFS) starting from the root node. Subsequently, \tap\ evaluates the performance of each strategy with a cost model and selects the strategy offering the best performance.



\subsection{Graph Reconstruction}\label{sec:pattern-routing}

After strategy validation, the best strategy is chosen from the candidates based on the cost model. \tap\ reconstructs the best strategy back into the computational graph form with the help of the ShardingPatterns. 
Each ShardingPattern gets materialized into operators by replacing the original subgraphs with the parallelized implementation specified by SRC.
After the parallelized graph is ready, it is passed to the training framework backend for execution.

\subsection{Communication-Based Cost Model}\label{sec:cost-model}
\begin{figure}
    \centering
    \includegraphics[width=\linewidth]{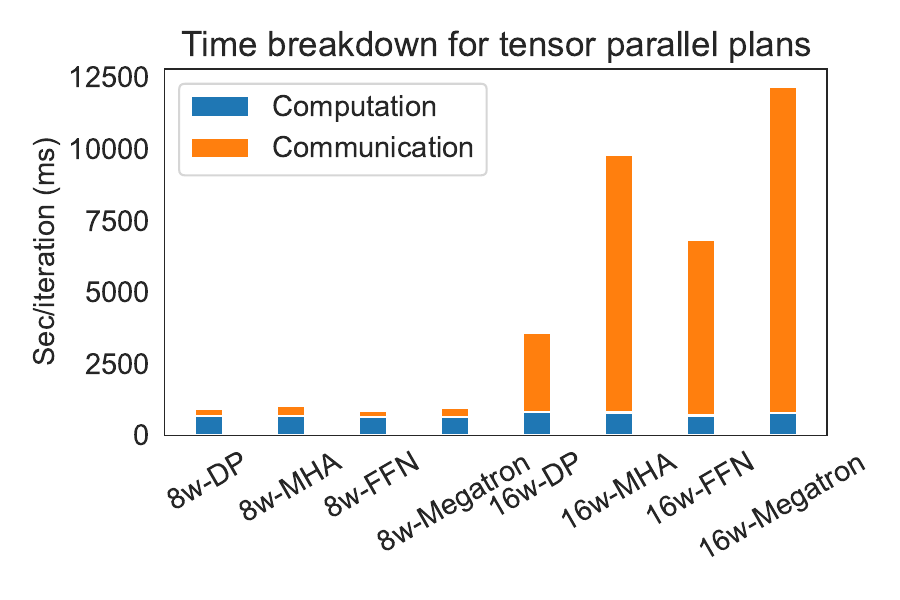}
    \caption{Profiling result for TP schedules of T5-large. \textit{DP}: data parallel, \textit{MHA}: sharding attention only, \textit{FFN}: sharding the feed-forward layer only. \textit{Megatron} refers to the strategy that shards both MHA and FFN.}
    \label{fig:time-breakdown}
\end{figure}

An accurate cost model is critical to the evaluation of candidate strategies. 
We profile different tensor parallel schedules of T5-large model using 8 and 16 GPUs (denoted as 8w/16w), and present the result in \autoref{fig:time-breakdown}. 

Each node is equipped with 8 GPUs and interconnected using Ethernet.
We observe that inter-node communication is the main bottleneck for tensor parallelism, because the internode interconnect (eg. Ethernet) is usually an order of magnitude slower than the intranode interconnect (eg. PCI-e or NVLink).
Therefore, \tap employs a communication-based cost model. 

Prior works~\cite{Unger2022UnityThe, Zheng2022Alpa:Learning, Jia2018BeyondNetworks,arzani2023rethinking} adopt the vanila $\alpha - \beta$ cost model, where $\alpha$ captures the network latency for each message, and $\beta$ captures the inverse of bandwidth. 
The total time to send a message of size $N$ is $T(N) = \alpha + \beta N$.

However, this does not account for the communication-computation overlapping scheme in deep learning frameworks. 
Specifically, we observe that overlapping exists in the backward propagation phase, inside the collective communication operation, and adjacent computation and communication.
We therefore propose two changes to the cost model to capture the overlapping effect on communication cost.


\paragraph{Gradient/weight update overlap in backward pass.} 
During the backward propagation phase, gradients are calculated with respect to model's parameter and synchronized across workers.
Instead of waiting for the entire gradient computation to finish before starting communication, most DL frameworks implement a gradient overlapping techniques allow the gradient to be communicated as soon as they are computed.
Therefore, gradient synchronization on later layers without blocking the backward computation on early layers. 
We find this optimization technique to reduce the total execution time of backward phase, and use a discount factor $\gamma$ ($0< \gamma \leq 1$) to quantify the extent of gradient overlap during the backward pass.

\paragraph{Communication-reduction overlap in collective communications.} 
Collective communication may be overlapped with preceding/succeeding communication operations by decomposition into smaller and data-independent steps~\cite{wang2022overlap, chen2024centauri, rashidi2021enabling}.
Inside the collective communication operation (eg. AllReduce), the reduction and communication may also be overlapped. 
\tap uses a coefficient $\epsilon$ ($0< \epsilon \leq 1$) to capture the overlapping effect of each collective communication primitive, collected through offline profiling.


\paragraph{Solution.}
\tap addresses these issues using an analytical cost model that treats backward/forward pass separately and different collectives separately.
The total cost is the summation of all costs of sharding patterns $p$ found along the computational graph's critical path.
The cost of each sharding pattern is determined by the latency and transmission delay, where the latency is linear with respect to the number of participating workers ($W$), and transmission delay is a factor of the message size in forward pass($N_{fwd}$) and backward pass ($N_{bwd}$), the bandwidth ($1/\beta$), and overlapping factor of collective communication ($\epsilon$). 

\begin{align}
N_p &= N_{fwd(p)} + N_{bwd(p)} \\
T_{latency(p)} &= \alpha'\cdot W \\
T_{trans(p)} & = \beta \cdot (N_{fwd(p)} + \gamma*N_{bwd(p)}) \cdot \epsilon \\
Cost (P_G, S) &= \sum_{p=1}^{P} (T_{latency(p)} + T_{trans(p)})
\end{align}





\section{Evaluation}

We seek to answer the following questions during the evaluation:
\begin{itemize}
    \item Search speed and training performance: How fast are the \textit{search} time and \textit{throughput} of \tap compared with other automatic and manual frameworks?
    \item Scaling experiments: How well can \tap\ scale on larger models and larger systems?
    \item Interpretation of discovered strategies: What can be learned from the discovered parallel strategies? 
    \item Micro benchmarks: How robust is the subgraph mining algorithm? 
\end{itemize}

\subsection{Evaluation Setup}
\tap is implemented using 12K lines of Python code on TensorFlow (TF). 
We assess \tap across a diverse set of models, including the dense transformer model (T5), the sparse mixture-of-experts model (GShard MoE), and the convolutional neural network (ResNet). T5 features an encoder-decoder transformer architecture, representing a broad spectrum of large language models like BERT, Llama-1/2/3, GPT-1/2/3, and M6. 

We choose Alpa as a evaluation baseline for automatic intra-op parallel framework. 
For expert-engineered tensor parallel frameworks, we employ Megatron ~\cite{Narayanan2021EfficientMegatron-LM} for the T5 model and DeepSpeed ~\cite{rasley2020deepspeed} for non-transformer models such as ResNet and GShard MoE, in line with the evaluation methods in ~\cite{Zheng2022Alpa:Learning, Unger2022UnityThe, cui2023optimizing}.

Given the diversity of backend frameworks in use, such as TensorFlow, PyTorch, and JAX, we follow the convention in ~\cite{Zheng2022Alpa:Learning, Narayanan2021EfficientMegatron-LM} by reporting the performance in FLOPs. 
This involves calculating the FLOPs for dense matrix multiplication operations per iteration and then dividing this by the iteration time. 
This method allows for a consistent and comparable evaluation of performance across different frameworks. 
In cases where evaluations are conducted on the same backend (between TAPAS and TensorFlow in \autoref{fig:scaling-perf}), we report performance metrics directly in terms of iteration time to ensure accuracy and relevancy. 

The evaluation was performed on Company A's public cloud nodes. 
Each node was equipped with 756GB main memory, $2\times$ Intel 8163 CPUs, and $8\times$ Nvidia V100 SXM2 32GB GPUs. 
The evaluations were performed using FP32 precision.
These compute nodes were interconnected by 100 Gbps ethernet.

Each representative model is scaled to various sizes. The T5 model is scaled by adding new layers (depth), the MoE model is scaled by adding experts and layers (width and depth), and the ResNet model is scaled by enlarging the classification layer (width).

\subsection{End-to-End Evaluation}\label{sec:e2e-search}
In this section, we compare with auto-parallel framework Alpa on both the search time and quality of the discovered strategies.

\subsubsection{Search time.}
\begin{figure}
\centering
    \includegraphics[width=0.9\linewidth]{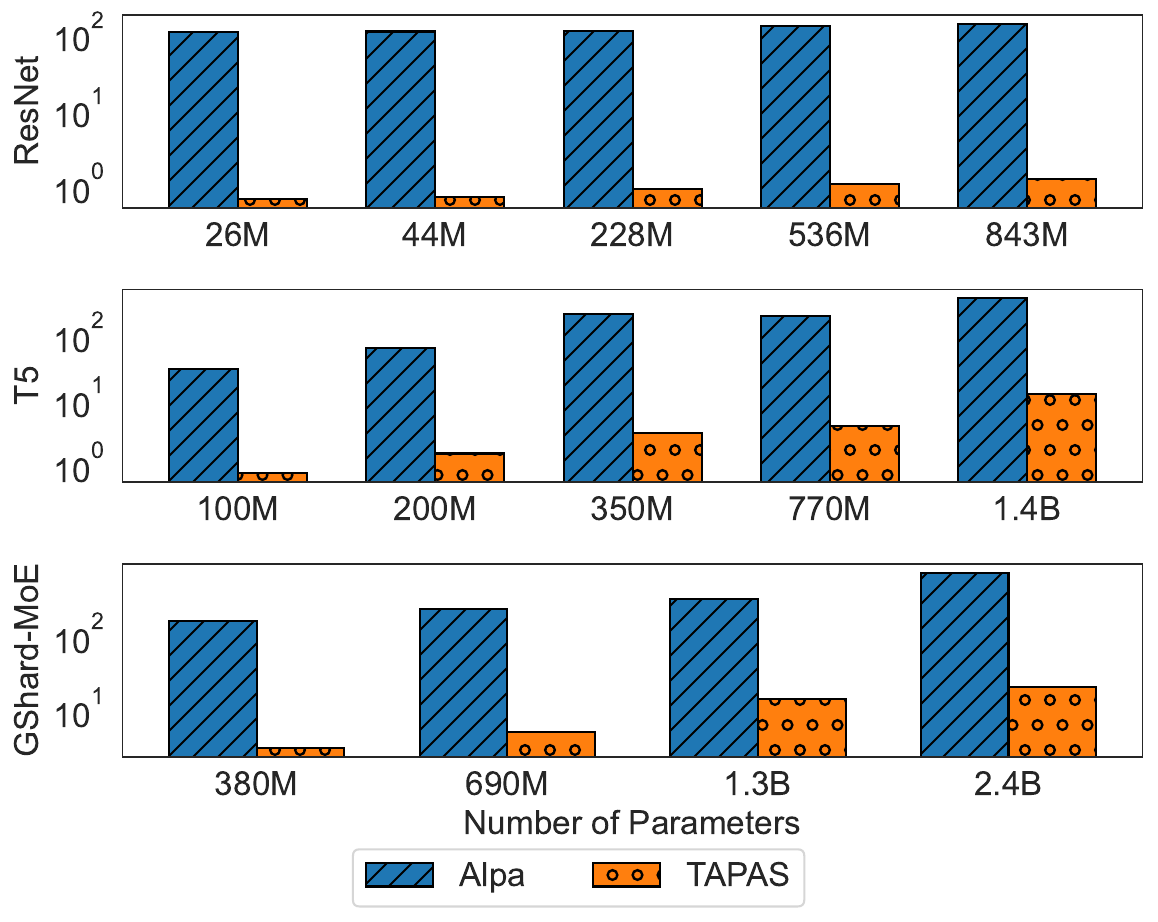}
    \caption{End-to-end search time (in minutes) under different frameworks.}
    \label{fig:scaling-params-search}
\end{figure}

Search time is defined by the duration of strategy derivation, excluding framework initialization time and actual training time. 
In \autoref{fig:scaling-params-search}, we present the end-to-end search time for increasing model sizes. 

To scale the model size along the width, we increase the size of the classification layer of the ResNet model. The base ResNet50 model has 1024 classes in the fully connected (FC) layer. As we increase the dimensions for the FC layer from 1024 to 10K, 100K, 250K, and 400K, the total number of parameters also scales up. As shown in the large-scale classification task with ResNet, \tap is two orders of magnitude faster than Alpa in finding the optimal solution, outperforming the latter by $103 - 162\times$.

To scale the model along the depth, we increase the number of transformer layers for T5. \autoref{fig:framework-flops-perf} shows that, with an increasing number of parameters, \tap can still find a plausible schedule in under 15 mins, which is $21 - 67\times$ faster than Alpa. 

We further analyze the time breakdown during the search. The efficient graph pruning algorithm greatly shrinks the search space while preserving key optimization space. Moreover, the analytical cost model used by \tap does not require operator profiling. As a result, Alpa takes 197 minutes to search 16 candidate strategies, while \tap requires only 6 minutes to examine 729 strategies for T5-large.

\subsubsection{Training speed.}
\begin{figure*}[t]
    \centering
\includegraphics[width=0.95\linewidth]{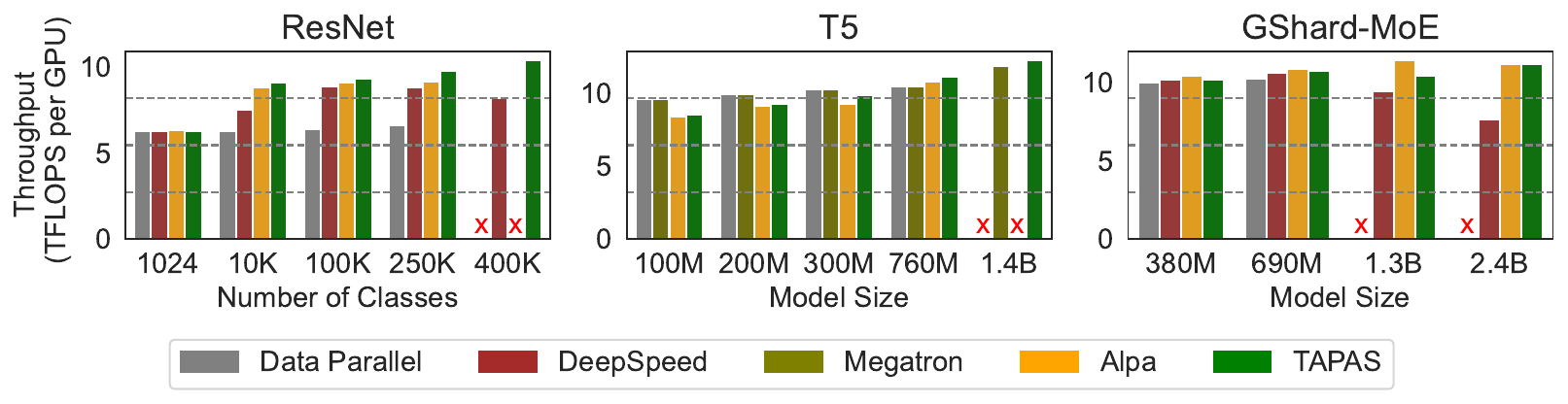}
    \caption{Performance across frameworks on 8 GPUs. "$\times$" represents out-of-memory failure. }
    \label{fig:framework-flops-perf}
\end{figure*}

\begin{figure*}[t]
  \centering
  \label{fig:scaling}
  \includegraphics[width=0.8\linewidth]{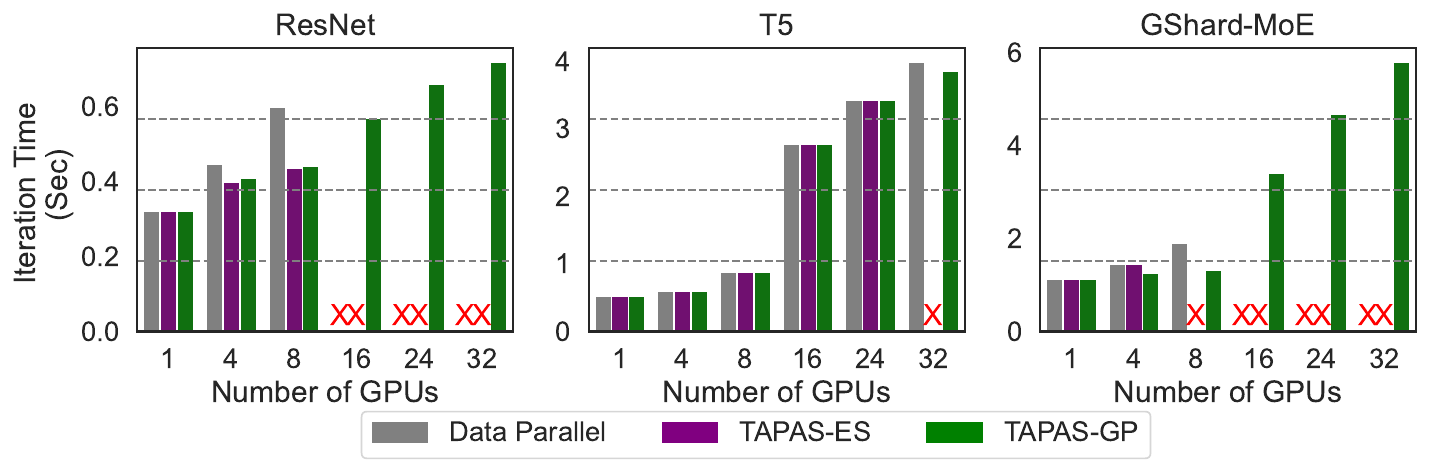}
    \caption{Weak scaling performance. TAPAS-ES: \tap with exhaustive search. TAPAS-GP: \tap with subgraph pruning.}
    \label{fig:scaling-perf}
\end{figure*}

We evaluate the performance of both manual and automatic parallel frameworks, and present them in \autoref{fig:framework-flops-perf}. 

\noindent\textbf{ResNet result.}
Both Alpa and \tap excel compared to data parallelism (DP), particularly in handling larger dense layers where Alpa tends to falter. 
DP duplicates the weight across all workers, quickly exhausting the memory. On the other side, DeepSpeed addresses the memory constraints by sharding the optimizer states and gradients across workers. However, this leads to an increase in the amount and size of messages, particularly for convolutional operators during the backward passes, impacting efficiency. \tap shards the fully connected (FC) layers while duplicating the base model, which effectively minimizes the memory burden and reduces the overhead associated with transmitting smaller messages.

\noindent\textbf{T5 result.}
Data parallelism and Megatron perform better than both Alpa and \tap when the model size is less than 760M parameters, while both Alpa and \tap outperform them on larger models. This is because Alpa performs reduce-scatter optimization to save communication, while \tap uncovers a novel parallel strategy that shards the feed-forward layer while replicating the self-attention layer inside a transformer. The self-attention layer is usually computationally intensive, but the feed-forward layer has two dense matrices. Therefore, sharding only the feed-forward layer can reduce the amount of weight updates while keeping the computational intensity high.

\noindent\textbf{GShard-MoE result.}
In the GShard-MoE experiments, \tap found an expert-level parallel strategy similar to Alpa, but it was discovered using a much smaller search space. The strategy partitions the expert dimension in the MoE layers, while it replicates the weight for non-expert layers like attention and MoE gates. DeepSpeed adopts a similar strategy as the original GShard implementation~\cite{Lepikhin2020GShard:Sharding}, and combines that with ZeRO-2 DP. However, DeepSpeed falls short when the number of experts does match with the number of worker GPUs. This highlights the need for automatic parallelism.

\subsection{Scaling Experiments}

We scale \tap on ResNet, T5, and MoE models with an increasing number of GPUs while keeping the per-GPU workload constant, and present the result in \autoref{fig:scaling-perf}.

\noindent\textbf{Baseline.}
The baseline is TensorFlow trained with data parallelism.
For all models, we first saturate the GPU memory by increasing the batch size until OOM occurs on a single GPU, and linearly scale the batch size with the number of GPUs. The size of the parameters of all base models (on 1 GPU) ranges from 0.77B to 1.3B.

\noindent\textbf{Result.}
We set a time limit of 120 minutes for the exhaustive search version of TAPAS. It is worth noting that the performance gap between the exhaustive search version of TAPAS (TAPAS-ES) and the subgraph-pruned version(TAPAS-GP) is within 1.5\% across the experiments. However, when increasing the size of compute cluster, TAPAS-ES frequently exceeds the time limit due to increased number of parallel strategies. 
This highlights the challenge of vast search space in auto-parallel frameworks. 

We also observe that \tap uncovers distinct sharding strategies on different cluster scales for the same model. 
For MoE (1.3B) model, \tap discovers that a nested Expert+Tensor Parallel strategy works best by further sharding the feedforward network within an expert layer. 
This can be helpful when using wide expert layers, which is common for language translation tasks.

On the ResNet (843M) model, we observe that when the size of the classification layer increases, more memory buffers are needed for caching gradients on each worker, hence the DP performance is worse than the \tap' schedule from 8 GPUs. 
After that, OOM error occurs on the DP schedule.

\subsection{Visualization of discovered strategies.}\label{sec:visual}
\begin{figure}
    \centering
\includegraphics[width=0.9\linewidth]{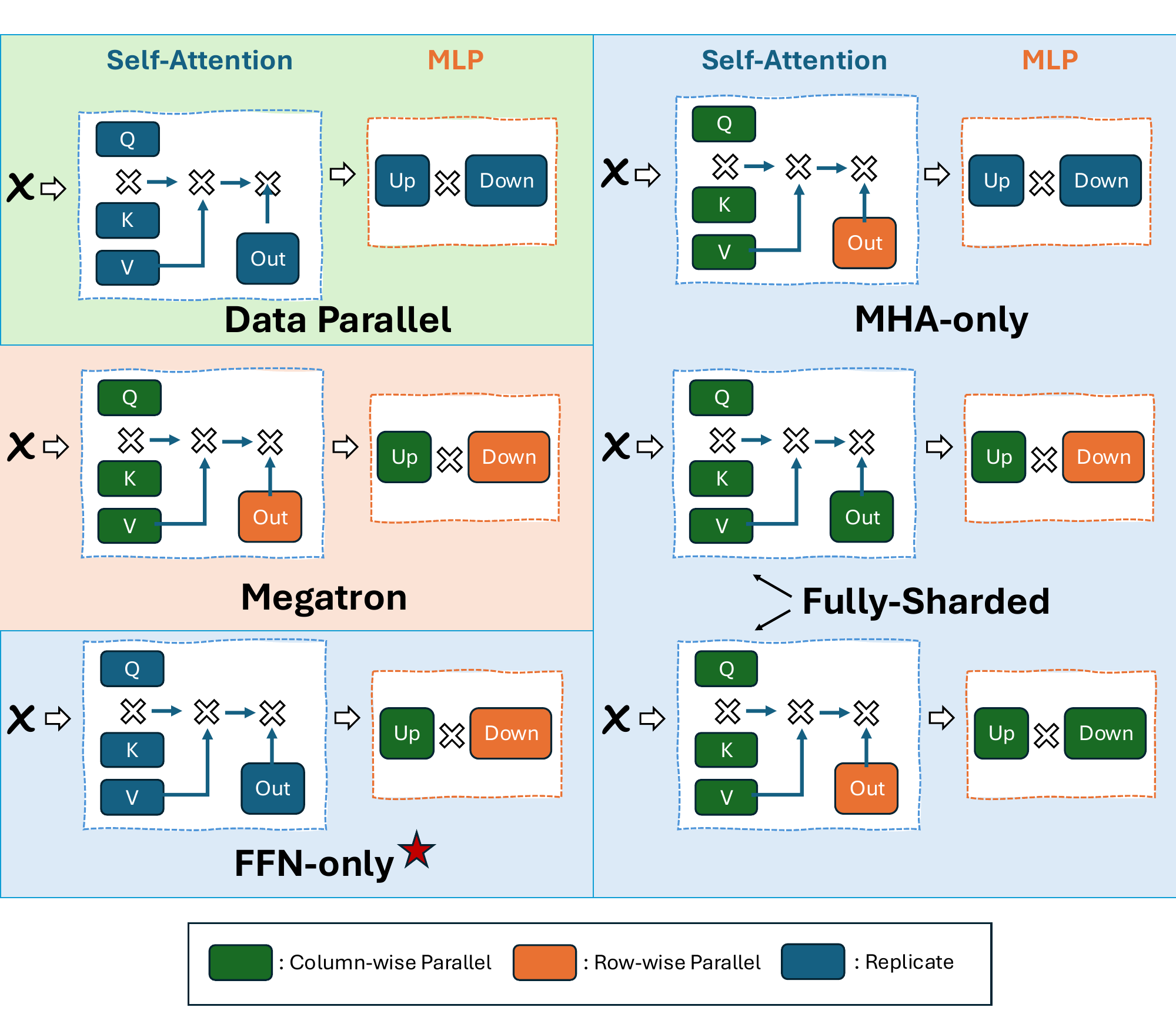}
    \caption{Visualization of selected sharding strategies discovered by \tap. LayerNorm/dropout/activation are ignored for presentation clarity. }
    \label{fig:sharding-plans}
\end{figure}

We plot the parallel strategies found by \tap in \autoref{fig:sharding-plans}. 
Our exploration uncovers that \tap is not only capable of identifying fully sharded strategies that mirror Megatron-LM, but also can it unearth novel strategies. 
These strategies involve partitioning either the multi-head attention (\textit{MHA-only}) or the feed-forward layers (\textit{FFN-only}). Intriguingly, the most effective strategy for dense transformers is the FFN-only plan. 
This is because FFN is composed of large MatMuls and can still achieve high arithmetic intensity even after splitting, whereas attention module has smaller weights and lower arithmetic intensity. 
Therefore, when memory ceiling permits, it is better to replicate the attention weight while splitting on the FFN weights. 
\tap also discovers other fully-sharded plans, but they are not selected as they will incur higher communication cost with the same amount of compute reduction. 
Our result challenges the belief that expert-engineered baseline is universally optimal, and demonstrates the effectiveness of our approach.

\subsection{Micro Benchmark}
\begin{figure}
    \centering
    \includegraphics[width=\linewidth]{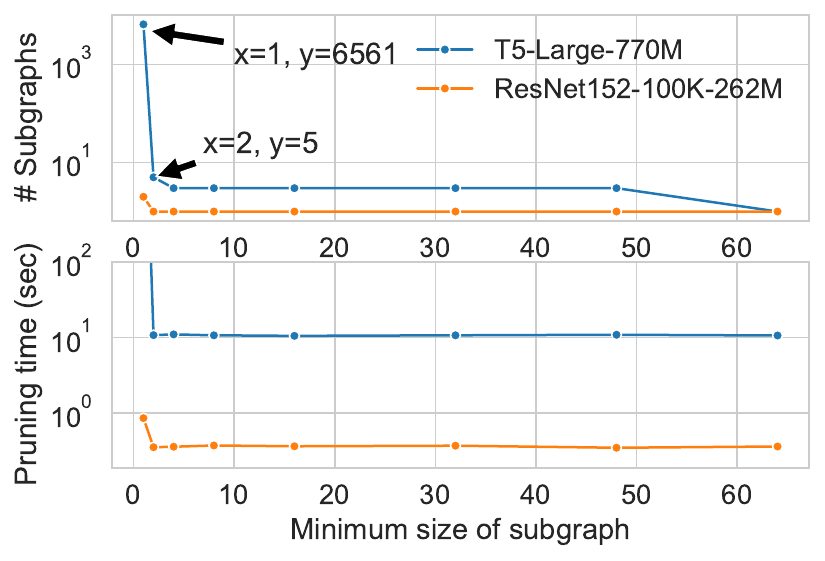}
    \caption{Subgraph pruning evaluations.}
    \label{fig:mb-mindup}
\end{figure}

In Algorithm 1, \texttt{minSize} determines the threshold for subgraph size. If the threshold is too low, it will still face the exploding search spaces by dealing with too many small graphs; if the threshold is too high, we may see too few subgraphs, resulting in a longer search time or sub-optimal policy. Since the architecture of different neural networks may vary significantly, it is desirable to have a robust threshold that does not require extensive tuning.

We explore a range of \texttt{minSize} and report the number of unique subgraphs found and subgraph mining algorithm running time in \autoref{fig:mb-mindup}. Take the T5-Large model with 770M parameters as an example: when the threshold is 1, meaning the graph is kept unfolded, it contains 6561 nodes. After subgraph mining, the number of unique subgraphs has drastically been reduced to just 5. As the threshold changes, the number of identified unique subgraphs stays relatively stable, showing that our algorithm is robust on different model architecture.

Furthermore, we observe that the subgraph mining algorithm is efficient, taking less than 12 seconds to find the subgraphs for T5-large, less than a second for the 152-layer 100K-class ResNet model, and 30 seconds for the 1.3B MoE network. This signals that \tap can scale well on large foundation models.

\subsection{Limitation and Future Work}
To extend \tap to pipeline parallel strategy, we can update the subgraph selection algorithm by choosing the sub-computation graphs as pipeline stages while satisfying load balancing constraints across subgraphs. 

To further optimize the memory consumption, \tap could leverage other orthogonal techniques such as mixed precision ~\cite{nvidiaamp}, gradient recomputation ~\cite{chen2016training, kirisame2021dynamic}.
Also, gradient checkpointing can be used to offload the selected GraphNode onto the main memory. 

It is worth noting that our approach may share some similarities with the pattern-matching technique used in TVM ~\cite{chen2018tvm} for operator fusion. 
The key difference is \tap targets the training setting, which is more challenging due to having dynamic tensor shape and unknown subgraph pattern.
TVM mainly targets the inference setting, where the patterns are pre-defined in the ML compilers based on expert experience, and the tensor shapes are known.
\section{Conclusion}
We present \tap, an automatic parallelism framework that efficiently discovers tensor parallel plans for large neural networks. Leveraging the observation that shared subgraphs widely exist in neural networks, we design TAPAS, an automatic parallel framework that significantly reduces redundant search effort by subgraph mining and early stopping. We also built an analytical cost model that accurately captures the amount of communication during tensor parallel training. The best parallel strategies discovered by \tap\ not only measure up to expertly engineered strategies, but also excel in search speed, reducing the strategy derivation time by two orders of magnitude compared to the state-of-the-art system. In summary, \tap\ provides a scalable, fast, and automatic solution for tensor parallelism that can help alleviate the burden of manual tuning.
\begin{acks}
This research is supported by the Ministry of Education, Singapore, under its Academic Research Fund Tier 1 (T1 251RES2409).
\end{acks}
\bibliographystyle{ACM-Reference-Format}

\appendix{}
\section{Complexity Analysis of Existing Works}\label{sec:complexity}
Following the discussion on related work, we analyze the complexities of two other automatic model parallel frameworks and present it in Table~\ref{tab:algo-complexities}. We define the total complexity as:

\begin{equation*}
\begin{split}
  total\_complexity = {}&search\_complexity \\
        &+ num\_plans*evaluation\_complexity\\
\end{split}
\end{equation*}

\subsubsection{FlexFlow.} FlexFlow operates on four dimensions (Sample, Operator, Attribute, and Parameter), and there was no space reduction. Therefore, the search space is $N (4E, 4V)$. As search complexity, FlexFlow employs the Markov chain Monte Carlo (MCMC) algorithm. Thus, we use $B$ to denote the computational budget (number of trials) in MCMC sampling. Furthermore, within each trial, it needs to evaluate its performance by querying the cost model with Depth-First-Search(DFS), hence its evaluation complexity is $O(V+E)$.

\subsubsection{Alpa.} Alpa is formulated as a multi-level optimization problem: in the outer loop, it searches for the inter-op parallel plan using dynamic programming; in the inner loop, it finds the intra-op parallel plan using integer linear programming. First, since it operates at MLIR HLO, which is a finer IR than the TensorFlow operator, we formulate the search space as $N(kE, kV)$ where $k \geq 1$. In the outer loops, it uses a similar algorithm to \cite{Li2021TeraPipe:Models} to search for pipeline slices and map the slices to devise mesh. Optimization like operator clustering and early pruning reduces the outer loop complexity to $(kV)^2L$. For the inner loop, since the exact complexity of their ILP solver is unknown, we use a lower bound by performing a BFS from each operator, and the complexity is given as $kE(kV+kE)$. Finally, each trial needs to evaluate its performance by querying the cost model, so the evaluation complexity is $kV+kE$.

\subsubsection{\tap} In \tap, we first reduce the search space by converting the TensorFlow graph to \tap graph (by $C \times$, where $C \geq 1$). We then prune the tree by layer, further reducing the complexity to $N(\frac{E}{2CL}, \frac{V}{2CL})$. In the searching stage, the result is derived by performing a BFS. Thus, the complexity is $\frac{V+E}{2CL}$. For the evaluation stage, \tap needs to evaluate the cost of each plan by querying the cost model, which depends only on the size of the edges. Thus, the evaluation cost is $\frac{E}{2CL}$. 

\section{Ablation Study on Cost Model Optimizations}
\tap cost model outputs a score for each candidate strategy and ranks them based on the score. We examine both the accuracy (whether the best strategy returned by the cost model is indeed the best strategy during runtime) and the order (how far off is the best strategy in the ranking). In this section, we study the effectiveness of optimizations on the cost model on 15 different architectures (5$\times$ T5, 6$\times$ CNN, and 4$\times$ MoE model).

For top-K candidates, we use the following metrics for evaluation:
\begin{itemize}
    \item Accuracy@K: the probability of the best-performing strategy being found within top-K of the cost model ranking; and
    \item Mean Reciprocal Ranking(MRR): the correctness of the ranking weighted by the relative position of the best strategy in ranking, defined as:
    \[MRR = \frac{1}{Q}\sum_{i=1}^{Q} \frac{1}{\text{rank}_i} \] 
    where $\text{rank}_i$ is the rank of the ground truth best strategy in the cost model ranking. 
\end{itemize}

MRR captures the relative order of strategies by encouraging better strategies placed at higher ranks. A higher MRR indicates that the best-performing strategy is ranked higher. Since we have 15 model architectures, $|Q|=15$. 

\begin{table}
\centering
  \caption{Ablation study of cost model optimizations. CF: constant filter, GO: Gradient Overlapping, EC: Efficiency of Collective Communications. }
  \label{tab:abl_study}
  \begin{tabular}{cccc|ccc}
    \hline
    Baseline & CF & GO & EC & Acc@1 &Acc@5 & MRR\\
    \hline
    \checkmark & &  &  & 0.53 & 0.86 & 0.71 \\
    \checkmark & \checkmark &  &  & 0.53 & 0.93 & 0.68 \\
    \checkmark & \checkmark & \checkmark &  & 0.73 & 1 & 0.84 \\
    \checkmark & \checkmark & \checkmark & \checkmark & 0.87 & 1 & 0.92\\
  \hline
\end{tabular}

\end{table}

From Table~\ref{tab:abl_study}, it's clear that the performance enhancements in our cost model primarily stem from communication overlap and the efficiency of collective communication primitives. The former optimization addresses non-overlapping communications, while the latter takes into account efficiency variances among primitives, resulting in better alignment with real training performance. The act of filtering constant tensors, while it may not yield significant improvements independently, is indeed a necessary step in the process.

\end{document}